\documentclass{ecai2008}
\usepackage{times}
\usepackage{graphicx}
\usepackage{latexsym}

\ecaisubmission   

\begin{document}

\title{Improved Statistical Machine Translation Using Monolingual Paraphrases}

\author{Preslav Nakov\institute{Linguistic Modeling Department
of the Institute for Parallel Processing
at the Bulgarian Academy of Sciences,
25A, Acad. G. Bonchev St., 1113 Sofia, Bulgaria,
email {\tt nakov@lml.bas.bg}}
\institute{Department of Mathematics and Informatics,
Sofia University,
5, James Bourchier blvd. 1164 Sofia, Bulgaria}
\footnote{Part of this research was done while the author was a PhD student
at the EECS department, CS division, University of California at Berkeley, USA.}}

\maketitle
\bibliographystyle{ecai2008}

\begin{abstract}
We propose a novel monolingual sentence paraphrasing method
for augmenting the training data
for statistical machine translation systems ``for free'' --
by creating it from data that is already available
rather than having to create more aligned data.
Starting with a syntactic tree, we recursively generate new sentence variants
where noun compounds are paraphrased using suitable prepositions,
and vice-versa -- preposition-containing noun phrases
are turned into noun compounds.
The evaluation 
shows an improvement equivalent to 33\%--50\% of that
of doubling the amount of training data.
\end{abstract}

\section{Introduction}

Most modern Statistical Machine Translation (SMT) systems rely on
aligned bilingual corpora (bi-texts) from which they
learn how to translate small pieces of text. In many cases,
these pieces are semantically equivalent
but syntactically different from translation-time text,
and thus the potential for high-quality translation can be missed.


In this paper, we describe a method for expanding
the training bi-text using paraphrases that are
nearly-equivalent semantically but different syntactically.
In particular, we apply sentence-level paraphrasing on the source-language side,
focusing on noun compounds (NCs) and noun phrases (NPs),
which have been reported to be very frequent in English written text:
2.6\% of the tokens in the {\it British National Corpus}
and 3.9\% in the {\it Reuters corpus}
are covered by NCs \cite{Baldwin:Tanaka:2004},
and about half of the words in news texts are part of an NP \cite{Koehn:Knight:2003:mt}.

The proposed approach is novel in that it augments the training corpus with
paraphrases of the original sentences, thus augmenting the training
bi-text without increasing the number of training translation pairs needed.
It is also monolingual; other related approaches map from the source
language to other languages in order to obtain paraphrases.
Finally, while our paraphrasing rules are English-specific,
the method is general enough to be domain-independent.



\section{Related Work}

Recent work in automatic corpus-based paraphrasing includes using
bi-texts as a source of alternative expressions for the same term
\cite{Barzilay:McKeown:2001:par,Pang:Knight:Marcu:2003:par},
or using multiple expressions of the same concept in one language \cite{Shinyama:al:2002:par}.
In a more recent work, \cite{Bannard:Callison-Burch:2005:par} propose
using phrases in a second language as pivots.
For example, if in a parallel English-German corpus, the English phrases
{\it under control} and {\it in check} happen to be aligned
(in different sentences) to the same German phrase {\it unter controlle},
they would be hypothesised to be paraphrases of each other with some
probability.

Recently, paraphrases have been used to improve machine translation {\it evaluation} .
For example, \cite{Kauchak:Barzilay:2006:par} argue that
automated evaluation measures like {\it Bleu} \cite{Papineni:Roukos:Ward:Zhu:2002}
end up comparing $n$-gram overlap
rather than semantic similarity with the reference text.
Having performed an experiment
asking two human translators to translate the same set of 10,000 sentences,
they found that less than .2\% of the translations were identical,
and 60\% differed by more than ten words.
Therefore, they proposed an evaluation method which paraphrases
the machine-produced translations and yields improved correlation
with human judgements compared to {\it Bleu}.
In a similar spirit, \cite{Zhou:al:2006:mt} use
a paraphrase table extracted from a bilingual corpus in order to improve
the evaluation of automatic summarization algorithms.

Another related research direction is in translating
units of text smaller than a sentence, e.g., NCs
\cite{Grefenstette:1999:web,Tanaka:Baldwin:2003:mt,Baldwin:Tanaka:2004},
NPs \cite{Cao:Li:2002:baseNP,Koehn:Knight:2003:mt},
named entities \cite{Al-Onaizan:Knight:2001:mt}, and technical terms
\cite{Nagata:al:2001:web}. While we focus on paraphrasing NPs/NCs,
unlike these approaches, we paraphrase and translate full sentences,
as opposed to working with small text units in isolation.

The approach we propose below is most closely related to that of
\cite{Callison-Burch:al:2006:mt}, who
translate English sentences into Spanish and French by
substituting unknown source phrases with suitable paraphrases.
Our paraphrases, however, are quite different.
Theirs are extracted with the above-mentioned bilingual method of
\cite{Bannard:Callison-Burch:2005:par}
using eight additional languages
from the {\it Europarl corpus} \cite{koehn:2005:europarl} as pivots.
These paraphrases are incorporated in the machine translation process
by adding them as additional entries in the phrase table and pairing
them with the foreign translation of the original phrase.
Finally, the system is tuned using minimum error rate training
\cite{Och:2003:mert} with an extra feature penalising
the low-probability paraphrases. This yielded dramatic increases in
coverage (from 48\% to 90\% of the test word types when 10,000 training
sentences were used), and notable increase on {\it Bleu} (up to 1.5\%).
However, the method requires large multi-lingual parallel
corpora, which makes it domain-dependent
and most likely limits its applicability to Chinese, Arabic, and the
languages of the EU, for which such large resources are likely to be available.

\section{Method}

We propose a novel general approach for improving SMT systems
using monolingual paraphrases.  Given a
sentence from the source (English) side of the training corpus,
we generate conservative meaning-preserving syntactic paraphrases
of that sentence.  Each paraphrase is paired with the foreign
(Spanish) translation that is associated with the original source sentence in
the training bi-text. This augmented training corpus is then used to
train an SMT system.

We further introduce a variation on this idea that can be used with a
\emph{phrase-based} SMT.  In this alternative, the source-language \emph{phrases}
from the phrase table are paraphrased,
but again using the target source-language phrase only,
as opposed to requiring a third parallel pivot language as in \cite{Callison-Burch:al:2006:mt}.
We also try to combine these approaches.



\section{Paraphrasing}
\label{sec:par}

Given a sentence like
``{\it I welcome the Commissioner's statement about the progressive and
  rapid lifting of the beef import ban.}'',
we parse it using the {\it Stanford parser} \cite{Klein:Manning:2003},
and we recursively apply the following syntactic transformations:

\begin{enumerate}
  \item {\bf [$_{\mathbf{NP}}$ NP$_1$ P NP$_2$] $\Rightarrow$ [$_{\mathbf{NP}}$ NP$_2$ NP$_1$].}\\
        {\it the lifting of the beef import ban} $\Rightarrow$
        {\it the beef import ban lifting}

  \item {\bf [$_{\mathbf{NP}}$ NP$_1$ \emph{of} NP$_2$] $\Rightarrow$ [$_{\mathbf{NP}}$ NP$_2$ gen NP$_1$].}\\
        {\it the lifting of the beef import ban} $\Rightarrow$
        {\it the beef import ban's lifting}

  \item {\bf NP$_{gen}$ $\Rightarrow$ NP.}\\
        {\it Commissioner's statement} $\Rightarrow$
        {\it Commissioner statement}

  \item {\bf NP$_{gen}$ $\Rightarrow$ NP$_{PP_{of}}$.}\\
        {\it Commissioner's statement} $\Rightarrow$
        {\it statement of (the) Commissioner}

  \item {\bf NP$_{NC}$ $\Rightarrow$ NP$_{gen}$.}\\
        {\it inquiry committee chairman} $\Rightarrow$
         {\it inquiry committee's chairman}

  \item {\bf NP$_{NC}$ $\Rightarrow$ NP$_{PP}$.}\\
        {\it the beef import ban} $\Rightarrow$
        {\it the ban on beef import}
\end{enumerate}

\noindent where:
  {\bf gen} is a genitive marker: ' or 's;
  {\bf P} is a preposition;
  {\bf NP$_{PP}$} is an NP with an internal PP-attachment;
  {\bf NP$_{PP_{of}}$} is an NP with an internal PP headed by {\it of};
  {\bf NP$_{gen}$} is an NP with an internal genitive marker;
  {\bf NP$_{NC}$} is an NP that is a noun compound.

The resulting paraphrases are shown in Table \ref{table:par}.
In order to prevent transformations (1) and (2) from constructing
awkward NPs, we impose certain limitations on NP$_1$ and NP$_2$.  They
cannot span a verb, a preposition or a quotation mark (although they can
contain some kinds of nested phrases, e.g., an ADJP in case of
coordinated adjectives, as in {\it the progressive and controlled
lifting}).  Therefore, the phrase {\it reduction in the taxation of labour}
is not transformed into {\it taxation of labour reduction} or {\it
taxation of labour's reduction}.
We further require the head to be a noun and we do not allow it to be
an indefinite pronoun like {\it anyone}, {\it everybody}, and {\it someone}.


Transformations (1) and (2) are more complex than they may look.
In order to be able to handle some hard cases,
we apply additional restrictions.
    First, some determiners, pre-determiners and possessive adjectives
must be eliminated in case of conflict between NP$_1$ and NP$_2$,
e.g., {\it \underline{the} lifting of \underline{this} ban}
can be paraphrased as {\it \underline{the} ban lifting},
but not as {\it \underline{this} ban's lifting}.
    Second, in case both NP$_1$ and NP$_2$ contain adjectives,
these adjectives have to be put in the right order,
e.g., {\it the \underline{first} statement of the \underline{new} commissioner}
can be paraphrased as {\it the \underline{first} \underline{new} commissioner's statement},
but not {\it the \underline{new} \underline{first} commissioner's statement}.
There is also the option of not re-ordering them, e.g.,
{\it the \underline{new} commissioner's \underline{first} statement}.
    Third, further complications are due to scope ambiguities
of modifiers of NP$_1$.  For example, in {\it the first statement of the new
commissioner}, the scope of the adjective {\it first} is not {\it statement} alone,
but {\it statement of the new commissioner}.  This is very different for the NP
{\it the biggest problem of the whole idea}, where the adjective {\it
biggest} applies to {\it problem} only, and therefore it cannot
be transformed to {\it the biggest whole idea's problem}
(although we do allow for {\it the whole idea's biggest problem}).

While the first four transformations are purely syntactic, (5) and (6)
are not.  The algorithm must determine whether a genitive marker is
feasible for (5) and must choose the correct preposition for (6). In
either case, for noun compounds of length three or more, we also need to
choose the correct position to modify,
e.g., {\it inquiry's committee chairman} vs. {\it inquiry committee's chairman}.

In order to improve the accuracy of the paraphrases,
we use the Web as a corpus, generating and testing the paraphrases
in the context of the preceding and the following words in the sentence.
First, we split the noun compound into two sub-parts
$N_1$ and $N_2$ in all possible ways, e.g., {\it beef import ban lifting} would be split as:
(a) $N_1$=``{\it beef}'', $N_2$=``{\it import ban lifting}'',
(b) $N_1$=``{\it beef import}'', $N_2$=``{\it ban lifting}'', and
(c) $N_1$=``{\it beef import ban}'', $N_2$=``{\it lifting}''.
For each split, we issue exact phrase queries
to {\it Google} using the following patterns:

\texttt{"lt $N_1$ gen $N_2$ rt"}

\texttt{"lt $N_2$ prep det $N_1'$ rt"}

\texttt{"lt $N_2$ that be det $N_1'$ rt"}

\texttt{"lt $N_2$ that be prep det $N_1'$ rt"}

\noindent where:
\texttt{$N_1'$} can be a singular or a plural form of $N_1$;
\texttt{lt} is the word preceding $N_1$ in the original sentence, if any;
\texttt{rt} is the word following $N_2$ in the original sentence, if any;
\texttt{gen} is a genitive marker ('s or ');
\texttt{that} is {\it that}, {\it which} or {\it who};
\texttt{be} is {\it is} or {\it are};
\texttt{det} is {\it the}, {\it a}, {\it an}, or none;
and \texttt{prep} is one of the prepositions
used by \cite{lauer:1995:thesis} for NC interpretation:
{\it about}, {\it at}, {\it for}, {\it from}, {\it in}, {\it of}, {\it on}, and {\it with}.

Given a particular split,
we find the number of page hits for each
instantiation of the above paraphrase patterns,
filtering out the ones whose page hit counts are less than ten.
We then calculate the total number of
page hits $H$ for all paraphrases (for all splits and all patterns),
and we retain the ones whose page hits counts are at least 10\% of
$H$, which allows for multiple paraphrases (possibly corresponding to different splits)
for a given noun compound.
If no paraphrases are retained, we repeat the above procedure with
\texttt{lt} set to the empty string.  If there are still no good
paraphrases, we set \texttt{rt} to the empty string.  If this does
not help either, we make a final attempt, by setting both \texttt{lt}
and \texttt{rt} to the empty string.
    For example, {\it EU budget} is paraphrased as {\it EU\underline{'s} budget} and
{\it budget \underline{of the} EU};
also {\it environment policy} becomes {\it policy \underline{on} environment},
{\it policy \underline{on the} environment},
and {\it policy \underline{for the} environment};
{\it UN initiatives} is paraphrased as {\it initiatives \underline{of the} UN},
{\it initiatives \underline{at the} UN}, and {\it initiatives \underline{in the} UN},
and {\it food labelling} becomes {\it labelling \underline{of} food}
and {\it labelling \underline{of} food\underline{s}}.

We apply the same algorithm to paraphrasing English {\it phrases} from
the phrase table, but without transformations (5) and (6).
See Table \ref{table:parPhraseTable} for sample paraphrases.

\begin{table}
  \centering
\newcommand\T{\rule{0pt}{2.6ex}}
\newcommand\B{\rule[-1.2ex]{0pt}{0pt}}
\begin{scriptsize}
\begin{tabular}{|ll|}
  \hline
{\bf 1} & {\bf \% of members of the irish parliament} \T \\
 & \% of irish parliament members\\
 & \% of irish parliament 's members\\
  \hline
{\bf 2} & {\bf universal service of quality .} \T \\
 & universal quality service .\\
 & quality universal service .\\
 & quality 's universal service .\\
  \hline
{\bf 3} & {\bf action at community level} \T \\
 & community level action\\
  \hline
{\bf 4} & {\bf , and the aptitude for communication and} \T \\
 & , and the communication aptitude and\\
  \hline
{\bf 5} & {\bf to the fall-out from chernobyl .} \T \\
 & to the chernobyl fall-out .\\
 \hline
{\bf 6} & {\bf flexibility in development - and quick} \T \\
 & development flexibility - and quick\\
 \hline
{\bf 7} & {\bf , however , the committee on transport} \T \\
 & , however , the transport committee\\
 \hline
{\bf 8} & {\bf and the danger of infection with aids} \T \\
 & and the danger of aids infection\\
 & and the aids infection danger\\
 & and the aids infection 's danger\\
 \hline
\end{tabular}
\end{scriptsize}
  \caption{\textbf{Sample English phrases from the phrase table
                    and corresponding automatically generated paraphrases.}}
  \label{table:parPhraseTable}
\end{table}


\begin{table*}
\begin{center}
\begin{tabular}{|l@{}|}
  \hline
    \textbf{\scriptsize I welcome the Commissioner 's statement about the progressive and rapid beef import ban lifting .}\\
    {\scriptsize I welcome the progressive and rapid beef import ban lifting Commissioner 's statement .}\\
    {\scriptsize I welcome the Commissioner 's statement about the beef import ban 's progressive and rapid lifting .}\\
    {\scriptsize I welcome the beef import ban 's progressive and rapid lifting Commissioner 's statement .}\\
    {\scriptsize I welcome the Commissioner 's statement about the progressive and rapid lifting of the {\it ban on beef imports} .}\\
    {\scriptsize I welcome the Commissioner statement about the progressive and rapid lifting of the beef import ban .}\\
    {\scriptsize I welcome the Commissioner statement about the progressive and rapid beef import ban lifting .}\\
    {\scriptsize I welcome the progressive and rapid beef import ban lifting Commissioner statement .}\\
    {\scriptsize I welcome the Commissioner statement about the beef import ban 's progressive and rapid lifting .}\\
    {\scriptsize I welcome the beef import ban 's progressive and rapid lifting Commissioner statement .}\\
    {\scriptsize I welcome the Commissioner statement about the progressive and rapid lifting of the {\it ban on beef imports} .}\\
    {\scriptsize I welcome the statement of Commissioner about the progressive and rapid lifting of the beef import ban .}\\
    {\scriptsize I welcome the statement of Commissioner about the progressive and rapid beef import ban lifting .}\\
    {\scriptsize I welcome the statement of Commissioner about the beef import ban 's progressive and rapid lifting .}\\
    {\scriptsize I welcome the statement of Commissioner about the progressive and rapid lifting of the {\it ban on beef imports} .}\\
    {\scriptsize I welcome the statement of the Commissioner about the progressive and rapid lifting of the beef import ban .}\\
    {\scriptsize I welcome the statement of the Commissioner about the progressive and rapid beef import ban lifting .}\\
    {\scriptsize I welcome the statement of the Commissioner about the beef import ban 's progressive and rapid lifting .}\\
    {\scriptsize I welcome the statement of the Commissioner about the progressive and rapid lifting of the {\it ban on beef imports} . }\\
\hline
\end{tabular}
    \caption{\textbf{Sample sentences and their automatically generated paraphrases.}
            Paraphrased noun compounds are in italics.}
    \label{table:par}
\end{center}
\end{table*}

\section{Experiments and Evaluation}

\subsection{Europarl Corpus}

We trained and evaluated several English$\rightarrow$Spanish phrase-based
statistical machine translation systems
using the {\it Europarl corpus} \cite{koehn:2005:europarl}
and the standard training/tuning/testing dataset splits.

First, we built English$\rightarrow$Spanish and Spanish$\rightarrow$English
directed word alignments using IBM model 4 \cite{Brown:al:1993:mt},
we combined them using the {\it intersect+grow heuristic} \cite{Och:Ney:2003:mt},
and we extracted phrase-level translation pairs
using the {\it alignment template approach} \cite{Och:Ney:2004:mt}.
We thus obtained a \emph{phrase table}
where each translation pair is associated with five parameters:
forward phrase translation probability, reverse phrase translation probability,
forward lexical translation probability, reverse lexical
translation probability, and phrase penalty.

We then trained a log-linear model using the following feature functions:
language model probability, word penalty, distortion cost, and the
above-mentioned parameters from the phrase table.
We set the feature weights by optimising the {\it Bleu} score directly
using {\it minimum error rate training} (MERT) \cite{Och:2003:mert}
on the first 500 sentences from the development set.
We then used these weights in a beam search decoder \cite{Moses:2007}
to translate the 2,000 test sentences,
and we compared the translations to the gold standard
using {\it Bleu} \cite{Papineni:Roukos:Ward:Zhu:2002}.

\textbf{Baseline.}
Using the above procedure, we built and evaluated a baseline system
$S$, trained on the original training corpus.

\textbf{Sentence-Level Paraphrasing.}
We further built $S_{pW}$, which uses a version of the training corpus
augmented with syntactic paraphrases of the \emph{sentences} from the English side
paired with the corresponding Spanish translations.
In order to see the effect of not breaking NCs and not using the Web,
we built $S_{p}$, which does not use transformations (5) and (6).

\textbf{Phrase Table Paraphrasing.}
System $S^\star$ paraphrases and augments the \emph{phrase table}
of the baseline system $S$ using syntactic transformations (1)-(4),
similarly to $S_{p}$, i.e., without NC paraphrasing.
Similarly, $S^\star_{pW}$ is obtained by paraphrasing the \emph{phrase tabl}e of $S_{pW}$.


\textbf{Combined Systems.}
Finally, we merged the phrase tables for some of the above systems,
which we designate with a ``$+$'', e.g., $S + S_{pW}$ and $S^\star + S^\star_{pW}$.
In these merges, the phrases from the first phrase table
are given priority over those from the second one
in case a phrase pair is present in both phrase tables.
This is important since the parameters estimated
from the original corpus are more reliable.

Following \cite{Bannard:Callison-Burch:2005:par}, we also performed an
experiment with an additional feature $F_{pW}$ for each
phrase: its value is 1 if the phrase is in the phrase table of $S$,
and 0.5 if it comes from the phrase table of $S_{pW}$.
As before, we optimised the weights using MERT.
For $S^\star + S_{pW}^\star$,
we also tried using two features: in addition to
$F_{pW}$, we introduced $F_{\star}$, whose value is 0.5 if the
phrase comes from paraphrasing a phrase table entry,
and 1 if it was in the original phrase table.


The evaluation results are shown in Tables \ref{table:eval10k} and \ref{table:eval}.
The differences between the baseline
and the remaining systems shown in Table \ref{table:eval10k} are statistically significant,
which was tested using bootstrapping \cite{Zhang:Vogel:2004:mt:conf}.

{\bf Gain of 33\%--50\% compared to doubling the training data.}
As Table \ref{table:eval} shows,
neither paraphrasing the training sentences, $S_{pW}$, nor
paraphrasing the phrase table, $S^\star$, lead to notable
improvements. For 10k training sentences,
the systems are comparable and improve {\it Bleu} by .3,
while for 40k sentences, $S^\star$ matches the baseline,
and $S_{pW}$ even drops below it.
However, merging the phrase tables of $S$ and $S_{pW}$,
yields an improvement of almost .7 for 10k and 20k sentences, and
about .3 for 40k sentences. While this improvement might look
small, it is comparable to that of \cite{Bannard:Callison-Burch:2005:par},
who achieved .7 improvement for 10k sentences,
and 1.0 for 20k (translating in the reverse direction: Spanish$\rightarrow$English).
Note also that the .7 improvement in {\it Bleu} for 10k and 20k sentences
is about 1/3 of the 2 {\it Bleu} point improvement achieved by the baseline system
by doubling the training size.
Note also that the .3 gain on {\it Bleu} for 40k sentences
is equal to half of what would have been gained if we had trained on 80k sentences.

{\bf Improved precision for all $n$-grams.}
Table \ref{table:eval10k} compares different systems trained on 10k sentences.
In addition to the {\it Bleu} score, we give its elements:
$n$-gram precisions, BP (brevity penalty), and ration.
Comparing the baseline with the last four systems,
we can see that all $n$-gram precisions are improved by about .4-.7 \emph{Bleu} points.

\begin{table}
\begin{tabular}{@{}l@{ }@{ }c|c@{ }c@{ }c@{ }c|c@{ }c|c@{ }c@{}}
  \hline
                             &  & \multicolumn{4}{c|}{\bf $n$-gram precision} &  \multicolumn{2}{c|}{\bf Bleu} & \multicolumn{2}{@{}c@{}}{\bf \# of phrases}\\
  {\bf System } & {\bf Bleu} & {\bf 1-gr.} & {\bf 2-gr.} & {\bf 3-gr.} & {\bf 4-gr.} & {\bf BP} & {\bf ration} & {\bf gener.} & {\bf used}\\
  \hline
  $S$ (baseline)             & 22.38 & 55.4 & 27.9 & 16.6 & 10.0 & 0.995 & 0.995 & 181k & 41k\\
  $S_{p}$                  & 21.89 & 55.7 & 27.8 & 16.5 & 10.0 & 0.973 & 0.973 & 193k & 42k\\
  $S_{pW}$                 & 22.57 & 55.1 & 27.8 & 16.7 & 10.2 & 1.000 & 1.000 & 202k & 43k\\
  $S^\star$                  & 22.58 & 55.4 & 28.0 & 16.7 & 10.1 & 1.000 & 1.001 & 207k & 41k\\
  $S + S_{p}$              & 22.73 & 55.8 & 28.3 & 16.9 & 10.3 & 0.994 & 0.994 & 262k & 54k\\
  $S + S_{pW}$             & {\bf 23.05} & 55.8 & 28.5 & 17.1 & 10.6 & 0.995 & 0.995 & 280k & 56k\\
  $S + S_{pW} \dag$    & {\bf 23.13} & 55.8 & 28.5 & 17.1 & 10.5 & 1.000 & 1.002 & 280k & 56k\\
  $S^\star + S_{pW}^\star$ & {\bf 23.09} & 56.1 & 28.7 & 17.2 & 10.6 & 0.993 & 0.993 & 327k & 56k\\
  $S^\star + S_{pW}^\star \ddagger$ & {\bf 23.09} & 55.8 & 28.4 & 17.1 & 10.5 & 1.000 & 1.001 & 327k & 56k\\
  \hline
\end{tabular}

  \caption{\textbf{\emph{Bleu} scores and $n$-gram precisions for 10k training sentences}.
           The last two columns show the total number of entries in the phrase table and
           the number of phrases that were usable at testing time, respectively.}
  \label{table:eval10k}
\end{table}

%

\begin{table}
  \centering
\begin{tabular}{|l|cccc|}
  \hline
                    & \multicolumn{4}{|c|}{\bf \# of training sentences}\\
  \multicolumn{1}{|c|}{\bf System}          & 10k   & 20k   & 40k & 80k\\
  \hline
  $S$ (baseline)   & \textbf{22.38} & \textbf{24.33} & \textbf{26.48} & \textbf{27.05}\\
  \cline{5-5}
  $S_{pW}$   & 22.57 & 24.41 & 25.96 & \multicolumn{1}{|c}{}\\
  $S^\star$  & 22.58 & 25.00   & 26.48 & \multicolumn{1}{|c}{}\\
  $S + S_{pW}$    & {\bf 23.05} & {\bf 25.01} & {\bf 26.75} & \multicolumn{1}{|c}{}\\
  \cline{1-4}
\end{tabular}
  \caption{\textbf{\emph{Bleu} scores for different number of training sentences.}}
  \label{table:eval}
\end{table}

{\bf Importance of noun compound splitting.}
$S_{p}$ is trained on the training corpus augmented with paraphrased
sentences, where the NC splitting rules (5) and (6) are not used.
We can see that the results for this system go below the baseline:
while there is a .3 gain on \emph{Bleu} on unigram precision,
bigram and trigram precision go down by about .1.
More importantly, BP decreases as well:
since the sentence-level paraphrases
(except for genitives, which are infrequent)
convert NPs into NCs, the resulting sentences are shorter,
and thus the translation model learns to generate shorter sentences.
This is different in $S_{pW}$,
where transformations (5) and (6) counter-weight (1)-(4), thus balancing BP to 1.
A somewhat different kind of argument applies to $S + S_{p}$,
which is worse than $S + S_{pW}$, but not because of BP.
In this case, there is no improvement for unigrams,
but a consistent .2-.3 drop for bigrams, trigrams and fourgrams.
The reason is shown in the last column of Table \ref{table:eval}:
omitting rules (5) and (6) results in fewer training sentences,
which means fewer phrases in the phrase table and
therefore fewer ones usable at translation time.

{\bf More usable phrases.}
The last two columns of Table
\ref{table:eval10k} show that, in general, having more phrases in the phrase table
implies more usable phrases at translation time.
A notable exception is $S^\star$,
whose phrase table is bigger than those of $S_{p}$ and $S_{pW}$,
but yields less utile phrases.
Therefore, we can conclude that the additional phrases extracted from paraphrased sentences
are more likely to be usable at test time
than the ones generated by paraphrasing the phrase table.

{\bf Paraphrasing sentences vs. paraphrasing the phrase table.}
As Tables \ref{table:eval10k} and \ref{table:eval} show,
paraphrasing the phrase table, as in $S^\star$ ({\it Bleu} score 22.58),
cannot compete against paraphrasing the training corpus
followed by merging the resulting phrase table with the phrase table
for the original corpus\footnote{Note that $S^\star$
does not use rules (5) and (6).
However, as $S+S_{p}$ shows, the claim holds even
if these rules are excluded when paraphrasing whole sentences:
the \emph{Bleu} score for $S+S_{p}$ is 22.73 vs. 22.58 for $S^\star$.},
as in $S+S_{pW}$ ({\it Bleu} score 23.05).
We also tried to paraphrase the phrase table of $S+S_{pW}$,
but the resulting system $S^\star+S^\star_{pW}$ yielded
little improvement: 23.09 {\it Bleu} score. Adding the two
extra features, $F_\star$ and $F_{pW}$, did not
yield improvements as well: $S^\star + S_{pW}^\star \ddagger$
achieved the same {\it Bleu} score as $S^\star+S^\star_{pW}$.
This shows that extracting additional phrases from the augmented corpus is
a better idea than paraphrasing the phrase table,
which can result in erroneous splitting of noun phrases.
Paraphrasing whole sentences as opposed to paraphrasing
the phrase table could potentially improve
the approach of \cite{Callison-Burch:al:2006:mt} as well: while low probability
and context dependency could be problematic,
a language model could help filter the bad sentences out.
Such filtering could potentially improve our results as well.
Finally, note that different paraphrasing strategies could be used
when paraphrasing phrases vs. sentences.
For example, paraphrasing the phrase table can be done more aggressively:
if an ungrammatical phrase is generated in the phrase table,
it would most likely have no negative effect on translation quality
since it would be unlikely to be observed at translation time.

{\bf Quality of the paraphrases and comparison to \cite{Callison-Burch:al:2006:mt}.}
An important difference between our syntactic paraphrasing
and the multi-lingual approach of \cite{Callison-Burch:al:2006:mt} is that
their paraphrases are only contextually synonymous
and often depart significantly from the original meaning.
As a result, they could not achieve improvements by simply augmenting the phrase table:
this introduced too much noise and yielded accuracy that was
below their baseline by 3-4 {\it Bleu} points.
In order to achieve an improvement, they had to introduce an extra feature
penalising the low probability paraphrases and promoting the
original phrase table entries. In contrast,
our paraphrases are meaning-preserving and less context-dependent.
For example, introducing feature $F_{pW}$
which penalises phrases coming from the paraphrased corpus
in system $S + S_{pW} \dag$ yielded a tiny improvement
on {\it Bleu} score (23.13 vs. 23.05), i.e., the phrases
extracted from our augmented corpus
are almost as good as the ones from the original corpus.
Finally, note that our paraphrasing method is
\emph{complementary} to that of \cite{Callison-Burch:al:2006:mt} and therefore
the two can be combined: the strength of our approach is in
improving the \emph{coverage of longer phrases} using syntactic paraphrases,
while the strength of theirs is in improving the \emph{vocabulary coverage}
with words extracted from additional corpora
(although they do get some gain from using longer phrases as well).

{\bf Paraphrasing the target side.}
We also tried paraphrasing the target language side,
i.e., translating into English,
which resulted in decreased performance.
This is not surprising: the set of available source phrases remains the same,
and a possible improvement could only come from producing
a more fluent translation, e.g., from transforming an NP with an internal PP
into an NC. However, unlike the original translations,
the extra ones are a priori less likely to be judged correct
since they were not observed on training.

\subsection{News Commentary \& Domain Adaptation}

We further applied the proposed paraphrasing method 
to domain adaptation using the data from the ACL'07 Workshop on SMT:
1.3M words (64k sentences) of \emph{News Commentary} data
and 32M words of \emph{Europarl} data. 
We used the standard training/tuning/testing splits,
and we tested on \emph{News Commentary} data.

This time we used two additional features with MERT
(indicated with the $\prec$ operation): 
for the original and for the augmented phrase table,
which allows extra weight to be given to phrases appearing in both.
With the default distance reordering,
for 10k sentences we had 28.88 Bleu for $S+S_{pW}$ vs. 28.07 for $S$,
and for 20k we had 30.65 vs. 30.34.
However, for 64k sentences, there was almost no difference: 32.77 vs. 32.73.
Using a different tokenizer and a lexicalized reordering model,
we got 32.09 vs. 32.34, i.e., the results were worse. 

However, as Table \ref{table:ourExperiments} shows, 
using a second language trained on \emph{Europarl},
we were able to improve \emph{Bleu} to 34.42 (for $S+S_{pW}$) from 33.99 (for $S$).
Using $S_{pW}$ lead to even bigger improvements (0.64 Bleu) 
when added to $S^{news} \prec S^{euro}$, 
where an additional phrase table from \emph{Europarl} was used.
See \cite{Nakov:2008:WMT} for further details.

\begin{table}
  \centering
\begin{tabular}{|l|c|c|}
  \hline
  & \multicolumn{2}{c|}{\bf Language Models}\\
  \textbf{Model} & \textbf{News Only} & \textbf{News+Euro}\\
  \hline
  $S^{news}$          & 32.27 & \textbf{33.99}\\
  $S^{news} \prec S_{pW}^{news}$ & 32.09 & \textbf{34.42}\\
  \hline
  $S^{news} \prec S^{euro}$              & & \textbf{34.05}\\
  $S^{news} \prec S_{pW}^{news} \prec S^{euro}$ & & 34.25\\
  $S^{news} \prec S^{euro} \prec S_{pW}^{news}$ & & \textbf{34.69}\\
  \hline
\end{tabular}
  \caption{Bleu scores on the \emph{News Commentary} data (64k sentences).}
  \label{table:ourExperiments}
\end{table}

\section{Problems and Limitations}

Error analysis has revealed that the major problems
for the proposed method are incorrect PP-attachments in the parse tree,
and, less frequently, wrong POS tags (e.g., JJ instead of NN).
Using a syntactic parser further limits the applicability of the approach
to languages for which such parsers are available.
In fact, for our purposes,
it might be enough to use a shallow parser or just a POS tagger.
This would cause problems with PP-attachment,
but these attachments are often assigned incorrectly by parsers anyway.
    The main target of our paraphrases are noun compounds --
we turn NPs into NCs and vice versa --
which limits the applicability of the approach to languages
where noun compounds are a frequent phenomenon,
e.g., Germanic, but not Romance or Slavic.
    From a practical viewpoint, an important limitation
    is that the size of the phrase table and/or of the training corpus increases,
which slows down both training and translation,
and limits the applicability to relatively small corpora for computational reasons.
    Last but not least, as Table \ref{table:eval} shows,
the improvements get smaller for bigger training corpora,
which suggests it becomes harder to generate useful paraphrases
that are not already in the corpus.

\section{Conclusion and Future Work}

We presented a novel domain-independent approach for improving
statistical machine translation by augmenting the training corpus
with monolingual source-language side paraphrases,
thus increasing the training data ``for free'',
by creating it from data that is already available
rather than having to create more aligned data.

While in our experiments we used phrase-based SMT,
any machine translation approach that learns from parallel corpora could
potentially benefit from the idea of syntactic corpus augmentation.
At present, our paraphrasing rules are English-specific,
but they could be easily adapted to other Germanic languages,
which make heavy use of noun compounds; the general idea of automatically
generating nearly equivalent source-side syntactic paraphrases
can in principle be applied to any language.
The current version of the method should be considered preliminary,
as it is limited to NPs; still, the results are already encouraging, and
the approach is worth considering when building MT systems from small corpora,
e.g., in case of resource-poor language pairs, in specific domains, etc.

Better use of the Web could be made for paraphrasing noun compounds
(e.g., using verbal paraphrases),
and other syntactic transformations could be tried
(e.g., adding/removing complementisers like {\it that}
and commas from nonmandatory positions).

Even more promising, but not that simple,
would be using a tree-to-tree syntax-based SMT system
and learning suitable syntactic transformations
that can make the source-language trees
structurally closer to the target-language ones.
For example, the English sentence
``{\it Remember the guy who you are \underline{with}!}''
would be transformed into
``{\it Remember the guy \underline{with} whom you are!}'',
whose word order is closer to the Spanish
``{\it !`Recuerda al individuo \underline{con} quien est\'{a}s!}'',
which might facilitate the translation process.

Finally, the process could be made
part of the decoding, which would eliminate the need of paraphrasing
the training corpus and might allow dynamically generating
paraphrases both for the phrase table entries and for the target sentence
that is being translated.

\ack
This research was supported in part by NSF DBI-0317510
and by FP7-REGPOT-2007-1 SISTER.

\bibliography{bibliography}


\end{document}